\documentclass{article}

\usepackage{PRIMEarxiv}
\usepackage{ulem}
\usepackage{color}
\usepackage[utf8]{inputenc} 
\usepackage[T1]{fontenc}    
\usepackage{hyperref}       
\usepackage{url}            
\usepackage{booktabs}       
\usepackage{amsfonts}       
\usepackage{nicefrac}       
\usepackage{microtype}      
\usepackage{lipsum}
\usepackage{fancyhdr}       
\usepackage{graphicx}       
\graphicspath{{media/}}     
\usepackage{amsmath}
\usepackage{amssymb}
\usepackage{verbatim}
\usepackage[ruled]{algorithm2e}

\newtheorem{lemma}{Lemma}

\title{An Efficient 1 Iteration Learning Algorithm for Gaussian Mixture Model And Gaussian Mixture Embedding For Neural Network
}

\author{
   Weiguo Lu\\
  Department of Mathematics\\
  University of Macau \\
  Macau\\
  \texttt{yc07476@um.edu.mo} \\
   \And
   Xuan Wu \thanks{Corresponding author}\\
   Department of Mathematics\\z
   University of Macau \\
   Macau\\
   \texttt{yc27937@um.edu.mo} \\
   \And
  Deng Ding \\
  Department of Mathematics\\
  University of Macau \\
  Macau\\
  \texttt{dding@um.edu.mo} \\
   \And
   Gangnan Yuan\\
   Great Bay Institute for Advanced Study \\
   Dongguan, 523000, Guangdong, China\\
   School of Mathematical Sciences \\
   University of Science and Technology of China\\
   Hefei, 230026, Anhui, China\\
\texttt{ygn94189@gmail.com} \\
}

\begin{document}
\maketitle

\begin{abstract}
We propose an Gaussian Mixture Model(GMM) learning algorithm, based on our previous work of GMM expansion idea. The new algorithm brings more robustness and simplicity than classic Expectation Maximization (EM) algorithm. It also improves the accuracy and only take 1 iteration for learning. We theoretically proof that this new algorithm is guarantee to converge regardless the parameters initialisation. We compare our GMM expansion method with classic probability layers in neural network leads to demonstrably better capability to overcome data uncertainty and inverse problem. Finally, we test GMM based generator which shows a potential to build further application that able to utilised distribution random sampling for stochastic variation as well as variation control.
\end{abstract}


\section{Introduction}

The inducement of events are varied and regularly alter throughout the real application process, which frequently results in different modes of data, i.e., a large shift in the mean value and covariance of the data. For linear regression modeling, it is standard procedure to separate the observable data into various components, and then to perform a weighted combination based on the posterior probability that the data belong to each component.

The work of a Belgian astronomer, mathematician, statistician, and sociologist named Quitelet \cite{que1846} suggested the potential of dissecting typical mixes into their component parts as early as 1846. Holmes \cite{gh1892} was another early scholar who emphasized mixing. He made the argument that using a single average to gauge the wealth disparity was insufficient, therefore he created the idea of population mixture.
Dating back 130 years, Pearson\cite{pear1894}, a famous biometrician, statistician and eugenicist, combined two normal probability density functions with different means and different variances to fit the crab data provided according to his observation of the biological data. This is the first major analysis involving mixture models.

The research and application of probability distribution began in the same period. Pioneering work in this area included the work of a generalized theory by Newcomb \cite{sn1886} and the work by Pearson\cite{pear1894}. After this, aside from some contributions by Jeffreys\cite{sj1932} and Rao\cite{cr1948}, the use of maximum likelihood methods (ML) to fit mixed models did not receive enough attention until the 1960s. Day and Wolfe \cite{nd1969,jw1970}have published a number of technical reports on iterative schemes for ML methods of mixed distribution fitting. Sadly, this approach has not yet been able to produce a formal iterative scheme, and its convergence has not yet been demonstrated.

Successfully, Dempster et al.\cite{ad1977} used their famous expectation maximization (EM) algorithm to formalize this iterative scheme in general, which solved the problem and established the convergence of ML solutions of mixed problems on a theoretical basis. The creation of EM algorithm catalyzes the application and extensive research of finite mixture models. It has developed for nearly 100 year, Stigler \cite{ss1986} provides an absorbing account of this early work on mixtures.

Later, New Combing proposed an iterative reweighting scheme, which can be seen as an application of the Expectation Maximization algorithm(EM). With development, many normal mixture model extensions show remarkable results. Moreover, many mixture models for different problems are widely used. Such as, exponential mixture model\cite{Ngu2016}, t-distribution expansion model\cite{baek2010} and many normal mixture model modification and improvement methods\cite{aran2015, bier2003, ju2012, song2005, ver2003, xie2012}.

Among them, the classical Gaussian mixture model is a weighted combination of multiple Gaussian distributions, which is widely used in complex and huge data. Because such models are not only suitable for a wide range of multimodal models, but also have the advantage of simplicity and speed in calculation. However, due to the complexity of image data, financial data or industrial data, the data obtained by observing them are often too large to directly classify the components accurately. Therefore, it is necessary to extract features from observation data in advance. Then more convincing results can be obtained through the division of multiple modes and weighted integration modeling in the hidden variable space.

In order to find the optimal solution of the GMM fit, the most classical practice is to employ the expectation maximization algorithm (EM) to compute estimates of these unknown parameters. According to the eigenvalue decomposition law, any full covariance matrix can be represented by its eigenvalues and eigenvectors. We can see that to convert the diagonal matrix to the full covariance matrix, we only need to construct the orthogonal matrix in it. So the EM algorithm can fit the GMM quickly and accurately. However, this algorithm still has problems that need to be overcome. First, the iterative nature of the EM algorithm makes it very dependent on the initial input values. The selection of the initial value has a great impact on the fitting results\cite{bier2003, blo2013, kwe2013, mck1994, pac2001, shi2017}, so the observation and analysis of the data before the iteration are highly demanding. Second, the EM algorithm is easy to fall into local optimal values \cite{ amendola2015maximum, jin2016local,srebro2007there}. According to its convex function gradient descent convergence\cite{abb2008, kon1992, kon1993}, the basic EM algorithm is easy to fall into local extreme values and has poor fault tolerance. Third, EM does not perform well on functions with extreme or mixed features. EM algorithm based on Gaussian mixture is not suitable for fitting sharp or even discontinuous functions. Therefore, we need to seek convergence calculation methods with wider adaptability.

With a high-fidelity output mixture model estimated from the input mixture model, the GMMs can be used for non-Gaussian uncertainty propagation. Determining the best number of components and weights is often discussed by scholars, although each Gaussian component can be easily propagated by moment matching methods. Terejanu et al.\cite{gt2008} showed that the weights of Gaussian components can be assumed to be invariant by nonlinear mapping if and only if their variance tends to infinity. They also proposed two optimization schemes based on quadratic programming for the weight update problem in nonlinear uncertainty propagation\cite{gt2011}. Within the next two or three years, Faubel et al. developed a Gaussian mixture filter that recursively splits each filter into a fixed number of unscented Kalman filters to deal with non-Gaussian noise\cite{ff2009}. Horwood et al.\cite{hj2011} established a segmentation scheme by comparing Gaussian moments obtained using Gauss-Hermite quadrature estimates. Similarly, Huber\cite{mh2011} designed a nonlinear filter to segment the Gaussian density based on the component weights and traces of the covariance matrix. Vishwajeet et al. proposed a Gaussian mixture uncertainty propagation scheme with weight updating and splitting based on the numerically calculated error of the Kol-mogorov equation\cite{kv2018}. Tuggle et al. proposed a filter scheme that assessed the nonlinearity by Kullback-Leibler divergence between two Gaussian distributions of first-order and second-order extended Kalman filters, and then split the distribution along characteristic directions that choose large\cite{kt2018} along both the Jacobian and the derivative of the directional variance.

In our previous work\cite{lu2023efficient}, we proposed a GMM model structure to expand arbitrary unknown densities which inspired by Fourier expansion. A simple learning method is introduced to learn GMMs. Our learning algorithm's advantage is that it is not likelihood-based, allowing us to avoid drawbacks like many critical points and the local minimum problem associated with likelihood-based methods. We extend our earlier work\cite{lu2023efficient} in this paper. A novel algorithm is created with a clearer mathematical justification. In our earlier work, we demonstrated that, in terms of probability, GMM density estimation could be as accurate as frequency distribution estimation. In this study, we offer a more substantial mathematical demonstration that GMM can estimate a large set of density even in high-dimensional situations like mixed density with multivariate facets. In other words, the GMM's performance in estimating arbitrary density could always be relied upon. While estimating density, it could be used as a generalizing technique. The performance of our approach is demonstrated by experiments in both 1 and 2 dimensional examples.

Additionally, we were able to map any density using independent normals and a multilayer perceptron (MLP) network using our technique. This is especially useful for stochastic processes, among many other applications. Diffusion models, for instance, use the diffusion process to transform latent variables into a Normal distribution before gradually reverting to the original latent distribution. We could gain more control of the latent domain using this straightforward strategy. An Decoder Network(autoencoder) is used to test the effectiveness of our embedding strategy.  Complete embedding technique is provided in section 5.

Section 2 provides preliminary which state the general notation and related theorems.  Section 3 provides our prove of generality of GMMs as density estimation. Section 4 provides our 1 iteration GMM learning algorithm with proof and experiments. Section 5  introduce our GMM embedding technique and neural network experiment. The final section provides a summary and conclusions.

\section{Preliminary}
\subsection{Gaussian Mixture Model(GMM)}
The classical Gaussian mixture model is widely used in clustering problems. Its essence is to represent the distribution density function of sample points with the weighted average sum of several Gaussian functions. For data obtained $x\in \mathbb{R}^D$, the weight of the $n$-th Gaussian $\pi_n$, mean vector $M_n\in \mathbb{R}^D$, covariance matrix $\Sigma_n\in{\mathbb{R}^{+}}^D\times {\mathbb{R}^{+}}^D$ and $N,D\in \mathbb{N}^+$,
\begin{equation}
    G(x)=\sum_{n=1}^{N}\pi_n \phi(x;M_n, \Sigma_n).
\end{equation}
Among them, $\phi$ is noted as the normal distribution.

According to the definition of GMM, it is essentially a probability density function. The integrals of the probability density function over its scope must sum to 1. The probability density function of the whole GMM is linearly added by the probability density functions of several Gaussian components, and the integral of the probability density function of each Gaussian component must be 1. Therefore, in order to make the probability density integral of the whole GMM equal to 1, each Gaussian component must be assigned a weight whose value is not greater than 1, and the sum of the weights is 1. In other words,
\begin{equation}
    \sum_{n=1}^{N}\pi_n =1.
\end{equation}

\subsection{EM algorithm}
In order to use the maximum a posteriori probability method to determine which Gaussian distribution each pixel belongs to, we need to estimate the unknown parameters in the model. Here we show how to use the expectation maximization method (EM) to get estimates for these unknown parameters.

Calculate the probability of the model $G(x )=P(x|\Theta)$, $P(x|\Theta)$ is the joint probability of the data $x$ given by the parameter $\Theta=(\theta,M,\Sigma) $ where $\theta=(\pi_1,\pi_2,\dots,\pi_N)$, $M=(M_1,M_2,\dots,M_N)$ and $\Sigma=(\Sigma_1,\Sigma_2,\dots,\Sigma_N)$. As for hidden data, $\gamma_{dn}$ is usually defined as
\[
  \text{$\gamma_{dn}$} =
  \begin{cases}
    \!\begin{aligned}
       & \text{1,} \end{aligned}           & \text{The d-th observation is derived from the n-th model,} \\
    \text{0,} & \text{otherwise.}
  \end{cases}
\]
So the hidden data of an observation $x_d$ is $\gamma_d=(\gamma_{d1},\gamma_{d2},\dots,\gamma_{dN})$. For $\gamma=(\gamma_{1},\gamma_{2},\dots,\gamma_{D})$, the full likelihood function is
\begin{equation}
    P(x,\gamma|\Theta)=\prod_{n=1}^{N}\prod_{d=1}^D\left[\pi_n\phi(x_d|M_n,\Sigma_n)\right]^{\gamma_{dn}}.
\end{equation}
However, it is very difficult to directly differentiate unknown parameters above. In general, we need to convert multiple multiplications in the likelihood function into multiple additions. Then the logarithm is equal to
\begin{equation}
\begin{aligned}
    \log (P(x,\gamma|\Theta))&=\log(\prod_{n=1}^{N}\prod_{d=1}^D\left[\pi_n\phi(x_d|M_n,\Sigma_n)\right]^{\gamma_{dn}})\\
    &=\sum_{n=1}^{N}(\sum_{d=1}^D\gamma_{dn}\log\pi_n+\sum_{d=1}^D\gamma_{dn}\log(\phi(x_d|M_n,\Sigma_n)).
\end{aligned}
\end{equation}
Because the natural logarithm function is strictly monotonically increasing, the logarithm likelihood function and the likelihood function can be maximized at the same time. Its extreme point is the estimated value of the unknown parameter in the Gaussian mixture model. Then the em algorithm can be iterated from this function.

E-step:
\begin{equation}
    E(\log (P(x,\gamma|\Theta))=\sum_{n=1}^{N}(\sum_{d=1}^D E (\gamma_{dn})\log\pi_n+\sum_{d=1}^D E (\gamma_{dn})\log(\phi(x_d|M_n,\Sigma_n))
\end{equation}
To make it more concise and clear that let
\begin{equation} \Gamma_{dn}=E(\gamma_{dn}|x_d,\Theta)=\frac{\pi_n\phi(x_d|M_n,\Sigma_n)}{\sum_{n=1}^{N}\pi_n\phi(x_d|M_n,\Sigma_n)},
\end{equation}
and
\begin{equation}
    e_{n}=\sum_{d=1}^{D}E(\gamma_{dn}).
\end{equation}
Then the step can be rewritten as
\begin{equation}
\begin{aligned}
    Q(\Theta,\Theta^i)&=\sum_{\gamma}P(\gamma|x,\Theta^i)\log (P(x,\gamma|\Theta))\\
    &=E_{\gamma}(\log (P(x,\gamma|\Theta)|x,\Theta^i)\\
    &=E_{\gamma|x,\Theta^i}(\log (P(x,\gamma|\Theta))\\
    &=\sum_{n=1}^{N}(e_{n}\log\pi_n+\sum_{d=1}^D E (\gamma_{dn})\log(\phi(x_d|M_n,\Sigma_n))
\end{aligned}
\end{equation}

M-step:
\begin{equation}
    \Theta^{i+1}=\arg \max_{\Theta}Q(\Theta,\Theta^i).
\end{equation}
Taking the derivative of $Q(\Theta,\Theta^i)$ yields the partial derivative of each unknown quantity, we can make its partial derivative equal to 0. Then
\begin{equation}\label{com}
    \begin{aligned}
    M_n^{new}&=\frac{\sum_{d=1}^{D}\Gamma_{dn}x_d}{\sum_{d=1}^{D}\Gamma_{dn}},\\
    \Sigma_n^{new}&=\frac{\sum_{d=1}^{D}\Gamma_{dn}(x_d-M_n)^{-1}\Sigma^{-1}(x_d-M_n)}{\sum_{d=1}^{D}\Gamma_{dn}},\\
    \pi_n^{new}&=\frac{\sum_{d=1}^{D}\Gamma_{dn}}{D},
    \end{aligned}
\end{equation}
Thus, given an initial value, we can iterate back and forth to find the value content. It is generally known that the EM algorithm has a convergence property. Actually, from the conditional probability , we can get:
\begin{equation}
    p(x|\theta)=\frac{p(x,\gamma|\theta)}{p(\gamma|x,\theta)},
\end{equation}
and
\begin{equation}
    \log p(x|\theta)=\log p(x,\gamma|\theta)-\log p(\gamma|x,\theta).
\end{equation}
As we set above about $Q(\theta,\theta^i)$, we can also let $H(\theta,\theta^i)=\sum_{\gamma}p(\gamma|x,\theta^i)\log p(\gamma|x,\theta)$, which lead to
\begin{equation}
\begin{aligned}
    Q(\theta,\theta^i)-H(\theta,\theta^i)&=\sum_{\gamma}p(\gamma|x,\theta^i)\log p(x,\gamma ,\theta)-\sum_{\gamma}p(\gamma|x,\theta^i)\log p(\gamma|x,\theta)\\
    &=\sum_{\gamma}p(\gamma|x,\theta^i)\log \frac{p(x,\gamma |\theta)}{p(\gamma|x,\theta)}\\
    &=\sum_{\gamma}p(\gamma|x,\theta^i)\log p(x|\theta)\\
    &=\log p(x|\theta).
\end{aligned}
\end{equation}
Then the iteration can be expressed as
\begin{equation}
\begin{aligned}
    \log p(\theta,\theta^{i+1})-\log p(\theta,\theta^i)&=Q(\theta^{i+1},\theta^{i})-H(\theta^{i+1},\theta^i)-(Q(\theta^i,\theta^i)-H(\theta^i,\theta^i))\\
    &=Q(\theta^{i+1},\theta^{i})-Q(\theta^{i},\theta^i)-(H(\theta^{i+1},\theta^i)-H(\theta^i,\theta^i)),
\end{aligned}
\end{equation}
where $Q(\theta^{i+1},\theta^{i})-Q(\theta^{i},\theta^i)>0$ and $$H(\theta^{i+1},\theta^i)-H(\theta^{i},\theta^i)=\sum_{\gamma}p(\gamma|x,\theta^{i+1})\log p(\gamma|x,\theta)-\sum_{\gamma}p(\gamma|x,\theta^i)\log p(\gamma|x,\theta)=\log 1=0.$$ So we can see that $p(x|\theta^{i+1})>p(x|\theta^{i})$. Combine with $p(x|\theta)$ has an upper bound $1$, we have a lemma follow:
\begin{lemma}
As $p(x|\theta^{i+1})>p(x|\theta^{i})$ and $p(x|\theta)$ is bounded, there is a supremum $A$ satisfying

(1)$ \forall  i\in N^+:\log p(x|\theta^i)\le A$.

(2)$\forall \varepsilon >0,\exists \log p(x|\theta^{i_0})>A-\varepsilon$.

For $i>i_0$, we have $A-\varepsilon <\log p(x|\theta^{i_0})\le \log p(x|\theta^i)\le A$ that $|\log p(x|\theta^i)-A|<\varepsilon $ and $$\lim_{i\rightarrow \infty }\log p(x|\theta^i)=A.$$
\end{lemma}
 That is the true that $\log p(x|\theta^i)$ converges to value $A$.

\subsection{GMM Expansion Algorithm}
We proposed a modeling concept called GMM expansion in our prior work.  We also develop a learning algorithm under our idea of GMM expansion\cite{lu2023efficient}. Fourier series has a major influence on the GMM expansion idea. Our underlying assumption is that every density may be approximately expanded by a Gaussian mixture. The component distributions can be seen as base frequencies in the Fourier expansion. This means that $\mu, \sigma$ of the component distributions are non-parametric and do not need to be optimized. In practice, we spread $\mu$ evenly across the dataset area. $\Sigma$ can be thought of as a hyper-parameter that modifies the general smoothness of the GMM density. In this case, $\pi$ is the only remaining parameter for us to optimize. It is no longer required to learn the optimum set of $\pi,\mu, \sigma$. We focus solely on finding the best set of $pi$ by learning from the dataset. Here we briefly explain the concept of GMM expansion and the learning algorithms proposed in \cite{lu2023efficient}:

\begin{itemize}
    \item $g\left(x\right)=\sum_n \pi_i \phi_i\left(x\right)$ is the density of GMM;
    \item $\sum_n \pi_i=1$, $\phi_i\left(x\right) \sim N\left(\mu_i,\sigma\right)$, $n$ is the numbers of normal distributions;
    \item $r=\frac{\text{max}\left(X\right)-\text{min}\left(X\right)}{n}$ is the interval for locating $\mu_i$;
    \item $\mu_i=\text{min}\left(X\right)+i\times r$ are means for Gaussian components;
    \item $\sigma= t\times r$ is variance for all Gaussian components;
    \item $t$ is a real number hyper-parameter, usually $1\leq t\leq 5$.
\end{itemize}

\begin{equation}\label{eq8}
    \pi^{+1}=\pi+\frac{\widetilde{dL\left(k_i^{mid},k_i^{side}\right)}}{d\pi}
\end{equation}
\begin{equation}\label{eq7}
     \frac{\widetilde{dL\left(k_i^{mid},k_i^{side}\right)}}{d\pi}=\begin{bmatrix}\frac{\text{d}\widetilde{Loss\left(x_j\right)}}{\text{d}\pi_0} \\ \vdots \\\frac{\text{d}\widetilde{Loss\left(x_j\right)}}{\text{d}\pi_n}  \end{bmatrix}=\begin{bmatrix}\widetilde{P_{f_0}}\left(k_i^{mid}\right)-\widetilde{P_{f_0}}\left(k_i^{side}\right)\\ \vdots \\\widetilde{P_{f_n}}\left(k_i^{mid}\right)-\widetilde{P_{f_n}}\left(k_i^{side}\right)\end{bmatrix}
\end{equation}
where $\widetilde{P_{g_i}}\left(k_i\right)=\int_{k_i}\phi_i\left(x\right)dx$, $\phi_i$ is the density of $i$th component distribution, $k_i^{mid}\in\left(\mu_i- d,\mu_i+ d\right]
$, and $k_i^{side}\in\left(\mu_i+r,\mu_i+r-d\right]$ or $k_i^{side}\in\left(\mu_i-r,\mu_i-r+d\right]$.

A simple version of our learning algorithm is provided in algorithm 1.

\begin{algorithm}
	\caption{Simplified GMM expansion Learning Algorithm }
	\label{alg:1}
	\BlankLine
	Initialization
	
	1.Define n Gaussian distributions and evenly spread $\mu$ across $Max\left(X\right),Min\left(X\right)$;
	
	2.Calculate r, $r=\frac{Max\left(X\right)-Min\left(X\right)}{n}$
	
	3.Define hyper parameter $\sigma$ respect to r, e.g., $\sigma=1r,\sigma=3r,$
	
	4.Define hyper parameter $d$ respect to $\sigma$, e.g., $d=\sigma /4,d=\sigma /6,$
	
	5.Initialize $\pi_1=\pi_2...=\pi_n=1/n$
	
	6.Define $k_i^{mid}\ and\ k_i^{side}$
	
	7.Approximate  $\widetilde{dL}_i\approx \widetilde{P_{f_i}}\left(k_i^{mid}\right)-\widetilde{P_{f_i}}\left(k_i^{side}\right)$.
	
	\ForEach{$x_j$}{
	    Find $\mu_i$ which closest to $x_j$;

		Update $\pi_i^{+1}=\pi_i+\widetilde{dL}_i$ ;
		
		for all m which $m \neq i$, $\pi_m^{+1}=\pi_m - \frac{\widetilde{dL}_i}{n}$;
	}
		
	Finally, re-normalize $\pi$s to satisfy $\sum \pi_i=1$
\end{algorithm}

In previous study\cite{lu2023efficient}, we didn't provide our learning method a theoretical basis in mathematics. In this work, we present a novel 1 iteration learning algorithm while adhering to the same GMM expansion model structure. The mathematical explanation is clearer than in earlier work. Convergence is proven and is more clearly stated and easy to apply in practice.

\section{1 iteration GMM expansion learning algorithm}

\subsection{Algorithm}

In this section we shows our 1 iteration learning algorithm for our GMM set up. We also proof that our algorithm is guarantee converge to a local minimum.

\begin{gather*}
    \theta^{(j)}=(\pi_{1}^{(j)},\pi_{2}^{(j)}...\pi_{n}^{(j)}),\\
    \text{From (1)-(3) we define Function } L(\theta),\\ L(\theta)=\sum_{n=1}^N\sum_{d=1}^DP(\gamma=n|\theta)P(x_d,\mu_n)
\end{gather*}
where $P(\gamma=n|\theta)=\pi_n$ and $P(x_d,\mu_n)=\phi(x_d|M_n,\Sigma_n)$. Because we do not update $M_n,and \Sigma$ is a hyper-parameter where constant for all Gaussians, the conditional probability $\phi(x_d|M_n,\Sigma_n)$ is equivalent to $\phi_n(x_d).$

\begin{gather*}
    \frac{\partial L(\theta)}{\partial \pi_n}=l_n=\sum_{d=1}^DP(x_d,\mu_n)=\sum_{d=1}^D\phi_n(x_d),\\
\end{gather*}

\begin{equation}\label{eq14}
    \pi_{n}^{(j+1)}=\frac{\pi_{n}^{(j)}+l_n}{\sum_{n=1}^N\pi_{n}^{(j)}+l_n}=\frac{\pi_{n}^{(j)}+l_n}{1+\sum_{n=1}^Nl_n},
\end{equation}
Where $j$ is the j step of the algorithm. This matches our previous finding in \cite{lu2023efficient}. When the $\mu$ and $\sigma$ are fixed in the GMM, the gradient of $\pi$ is nearly constant with respect to the dataset for many cost functions, including likelihood and the suggested function $L(\theta)$.

\begin{algorithm}[ht]
	\caption{Proposed Learning Method}
	\label{alg:2}
	\BlankLine
	Initialization
	
	1.Define n Gaussian distributions and evenly spread $\mu$ across $\max\left(X\right),\min\left(X\right)$;
	
	2.Calculate r, $r=\frac{\max\left(X\right)-\min\left(X\right)}{n}$
	
	3.Define hyper parameter $\sigma$ respect to r, e.g., $\sigma=1r,\sigma=3r,$. If in higher dimensional cases, covariance matrix $\Sigma$ set to be identity matrix * $\sigma$.
	
	4.Initialize $\pi_1=\pi_2...=\pi_n=1/n$

        5.Calculate (\ref{eq14})
        
        6.Finally, re-normalize $\pi$s to satisfy $\sum \pi_i=1$ 
\end{algorithm}

\subsection{Theoretical Analysis}

The following Lemma captures the relationship of $\pi$ and data base on the idea of GMM expansion.

\begin{lemma}

Assume the following condition are satisfied:
(1) Define $N$ Gaussian Unit $\phi_1,...\phi_N$ and each $\phi_n\sim N(\mu_n,\Sigma)$. 

(2) Observed Dataset X, and $r=(\max(X)-\min(X))/N,\mu_n=\min(X)+r*n$. $\Sigma\leq\alpha$ which the same for all $\phi_n$ and small enough that satisfy $\phi_n(x=\mu_n)>1.0$. 

(3)Define $N$ $\pi_n$ where $\pi_1^{(0)}=\pi_2^{(0)}...=\pi_N^{(0)}=1/N$. Denote that $[\pi_1^{(j)}...,\pi_n^{(j)}]=\theta^{j}$. Gaussian Mixture distribution density function denoted as $G(x|\theta)$

(4)Under condition (1)-(3), learned $\pi$ from EM algorithm at the first step is denoted as $\pi_{0}^{EM(1)}...\pi_{n}^{EM(1)}$

(5)$\pi_{n}^{(j)}=\frac{\sum_{d=1}^D\phi_n(x_d)}{\sum_{n=1}^N\sum_{d=1}^D\phi_n(x_d)}$

The following equation hold:
\begin{equation}
    \pi_{n}^{(j)}-\pi_{n}^{EM(1)}<\epsilon
\end{equation}

With Lemma.1, we have:

\begin{equation}    \sum_{x=d}^D\log(G(x|\theta^{(j)}))>\sum_{x=d}^D\log(G(x|\theta^{(0)}) )
\end{equation}

\end{lemma}

\subsubsection{The proof of Lemma}
At (\ref{eq14}), we can derive that:
\begin{equation}\label{eq15}
    \pi_{n,j+1}=\frac{\pi_{n,j}}{1+\sum_{n=1}^Nl_n}+\frac{l_n}{1+\sum_{n=1}^Nl_n},
\end{equation}

Because we initialize $\mu_n$ across the data space evenly, when the number of Gaussian units is large and the variance is small, it is reasonable to assume that:

1. For any x in dataset, we can find at least 1 Gaussian unit with a mean close to that data point. 

2. When $\sigma$ decreases, $\phi(x,\mu,\sigma)$ increases. Together with assumption 1, we can assume that for every data point, 
\begin{equation}
l_n=\sum_{d=1}^DP(x_d,\mu_n)=\sum_{d=1}^D\phi_n(x_d)>1.0,
\end{equation}
(e.g.$\phi(x=0,\mu=0,\sigma=0.35)=1.329$.)

Together with (\ref{eq14}) and (\ref{eq15}), we can conclude that:
\begin{equation}
    \frac{l_n}{1+\sum_{n=1}^Nl_n}\leq\pi_{n,j+1}\leq\frac{1}{1+N}+\frac{l_n}{1+\sum_{n=1}^Nl_n}
\end{equation}

When N is large, our algorithm become less and less sensitive to the initial values of $\pi$, and approximately equal to $\frac{l_n}{1+\sum_{n=1}^Nl_n}$. Under this condition, our method is a one iteration method where we can perform (\ref{eq14}) for update or calculate:

\begin{equation}\label{eq16}
    \pi_{n}^{Our(j+1)}\approx\frac{l_n}{\sum_{n=1}^Nl_n}
\end{equation}

It is also important to discuss how well this estimation algorithm perform. By taking the same approach as EM algorithm, we can proof that we also share the same theoretical guarantees as EM algorithm does. 

Notice that in EM algorithm:
\begin{equation}
    Q^*(\theta,\theta^{i-1})=E(\log(W(x,\gamma|\theta))=\sum_{\gamma}p(\gamma|x,\theta^{i-1})\log(\gamma_i)
\end{equation}

\begin{equation}
P(\gamma=n|x_d,\theta)=\frac{\pi_n\phi_n(x_d)}{\sum_{n=1}^{N}\pi_n\phi_n(x_d)}=\frac{\pi_n\phi_n(x_d)}{G(x_d,\theta)}
\end{equation}

\begin{equation}
P(\gamma=n|X,\theta)=\sum_{d=1}^D\frac{\pi_n\phi_n(x_d)}{G(x_d,\theta)}
\end{equation}

\begin{figure}[ht]
	\centering
		\includegraphics[scale=0.4]{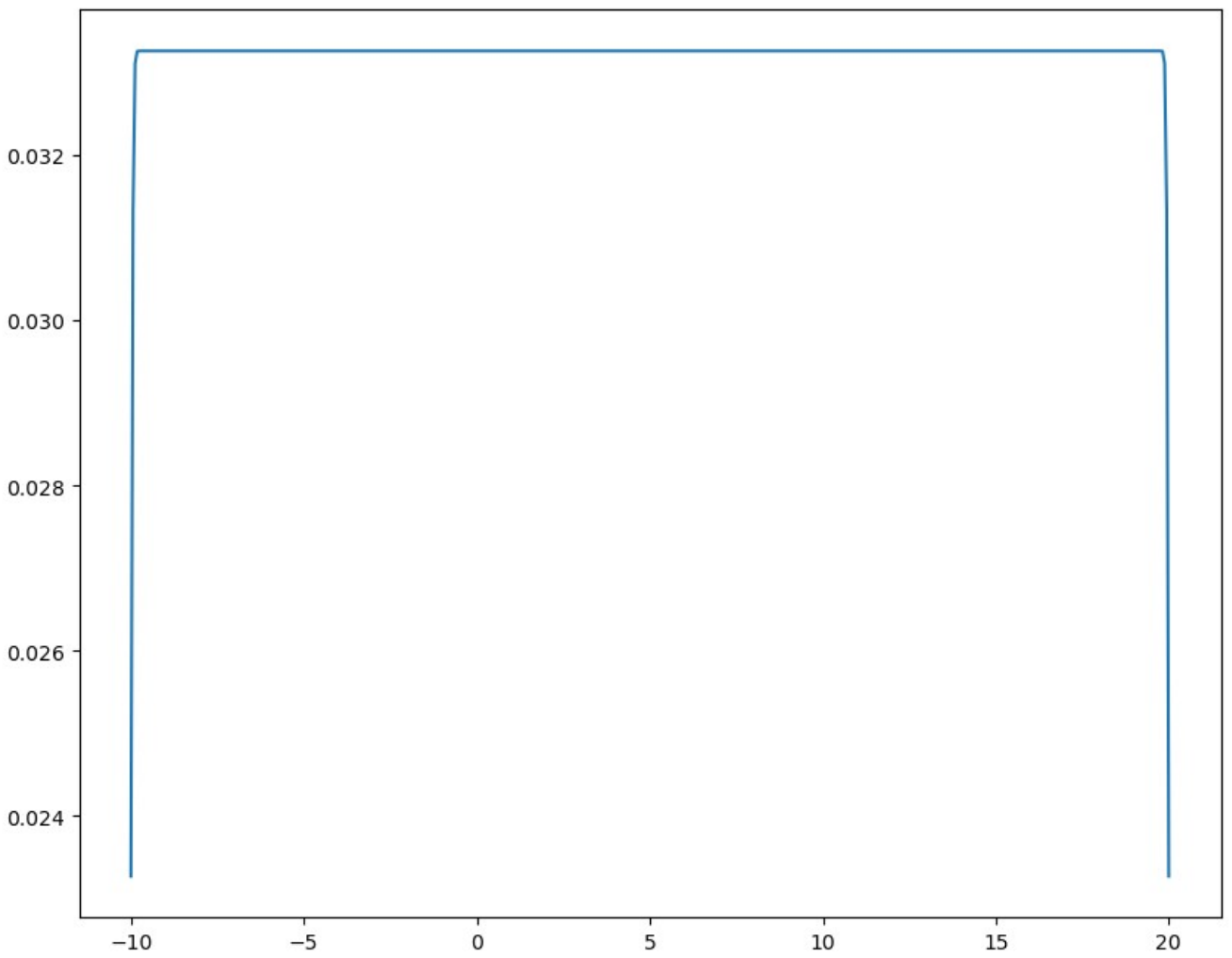}
	\caption{$G(x|\theta)$ with equally initialised $\theta$}
	\label{FIG:2}
\end{figure}

Because we equally initialise $\pi$ and evenly place the $\mu$ across the data space, the density values of initial GMM almost the same across the data domain which shows in Fig.\ref{FIG:2}. It is reasonable to derive that:

\begin{gather*}
 \pi_1^{(0)}=\pi_2^{(0)}...\pi_N^{(0)}=1/N,\\
G(x_1,\theta^{(0)})\approx G(x_2,\theta^{(0)})\approx...G(x_D,\theta^{(0)})=k,
\end{gather*}

\begin{equation}\label{eq17}
P(\gamma=n|X,\theta^{0})=\frac{\pi_n^{(0)}}{G(x_d|\theta^{(0)})}\sum_{d=1}^D\phi_n(x_d)=\frac{\pi_n^{(0)}}{G(x_d,\theta^{(0)})}l_n,
\end{equation}

Base on EM algorithm, we have following property where:
\begin{equation}\label{eq18}
    \pi_n^{EM(j)}=\frac{1}{N}P(\gamma=n|X,\theta^{(j-1)})=\arg \max_{\theta}(Q(\theta,\theta^{j-1}))
\end{equation}

Put (\ref{eq18}) into (\ref{eq17}) along with the initial $\pi$ and $G(x,\theta)$, we derive:

\begin{equation}
    \pi_{n}^{EM(1)}=N\frac{\pi_{n}^{(0)}}{G(x_d,\theta^{(0)})}l_n \approx \frac{1}{k}l_n
\end{equation}

\begin{equation}
    \frac{\pi_{n}^{EM(1)}}{\sum_{n=1}^N\pi_{n}^{EM(1)}}=\frac{\frac{1}{k}l_n}{\sum_{n=1}^N\frac{1}{k}l_n}=\frac{l_n}{\sum_{n=1}^Nl_n}=\pi_{n}^{EM(1)}
\end{equation}

From (\ref{eq16}), we conclude that:

\begin{equation}
    \pi_{n}^{Our(j+1)}\approx\pi_{n}^{EM(1)}=\frac{l_n}{\sum_{n=1}^Nl_n}
\end{equation}  
The proof is completed.

This proof shows that our estimation shares the same theoretical guarantee that the likelihood of learning GMM guarantees a better result $\log(G(X|\theta_1))\geq \log(G(X|\theta_0))$. In the next section, we use numerical experiments to compare our algorithm's efficiency with the EM algorithm. 

\section{Numerical Experiments}

In this section, We ran a number of tests to validate our suggested approach for learning GMM parameters and compare it with the conventional EM algorithm. Experiments in both one and two dimensions have been carried out.

\subsection{Metrics-Interval Probability Error}
Instead of KL-divergence or Wasserstein distance, we use a simple metric that we call interval probability error (IPE) to measure the difference between two distributions. We split the data space equally into $n$ intervals denoted as $\omega_i$. The purpose of this metric is to give a straight-forward and yet easy-to-calculate method to measure the difference between two densities. 

\begin{equation}
\text{IPE}=\sum_{i=1}^n\mid \int_{\omega_i}f\left(x\right)dx-\int_{\omega_i}g\left(x\right)dx \mid=\sum_{i=1}^n\mid P_f\left\{\omega_i\right\}-P_g\left\{\omega_i\right\}\mid,
\end{equation} 
where $f(x)$ is our target distribution and $g(x)$ is the approximate distribution. In our case, $g(x)$ is the Gaussian mixture distribution.

Density is rarely what we are ultimately after in most real-world scenarios. Probability is more comprehensive than density in comparison. IPE is arguably a better metric to assess the accuracy of our approximation in terms of probability than density. It has several benefits to calculate IPE instead of KL-divergence or Wasserstein distance. 1. It is easy to calculate. 2. The fact that this function is bound by $0\leq IPE\leq2$. The accuracy of approximation can be easily evaluated.

\subsection{Test Results}
We will demonstrate the differences between the suggested one-iteration approach and the EM algorithm in this section. Results from experiments in both one and two dimensions are illustrated.

\subsubsection{Graphical Comparison}

A density plot comparison of the estimation results allows access to the overall geometry and briefly illustrates the estimation accuracy. Both one and two-dimensional cases are shown as follows.

\begin{figure}[ht]
	\centering
		\includegraphics[scale=0.38]{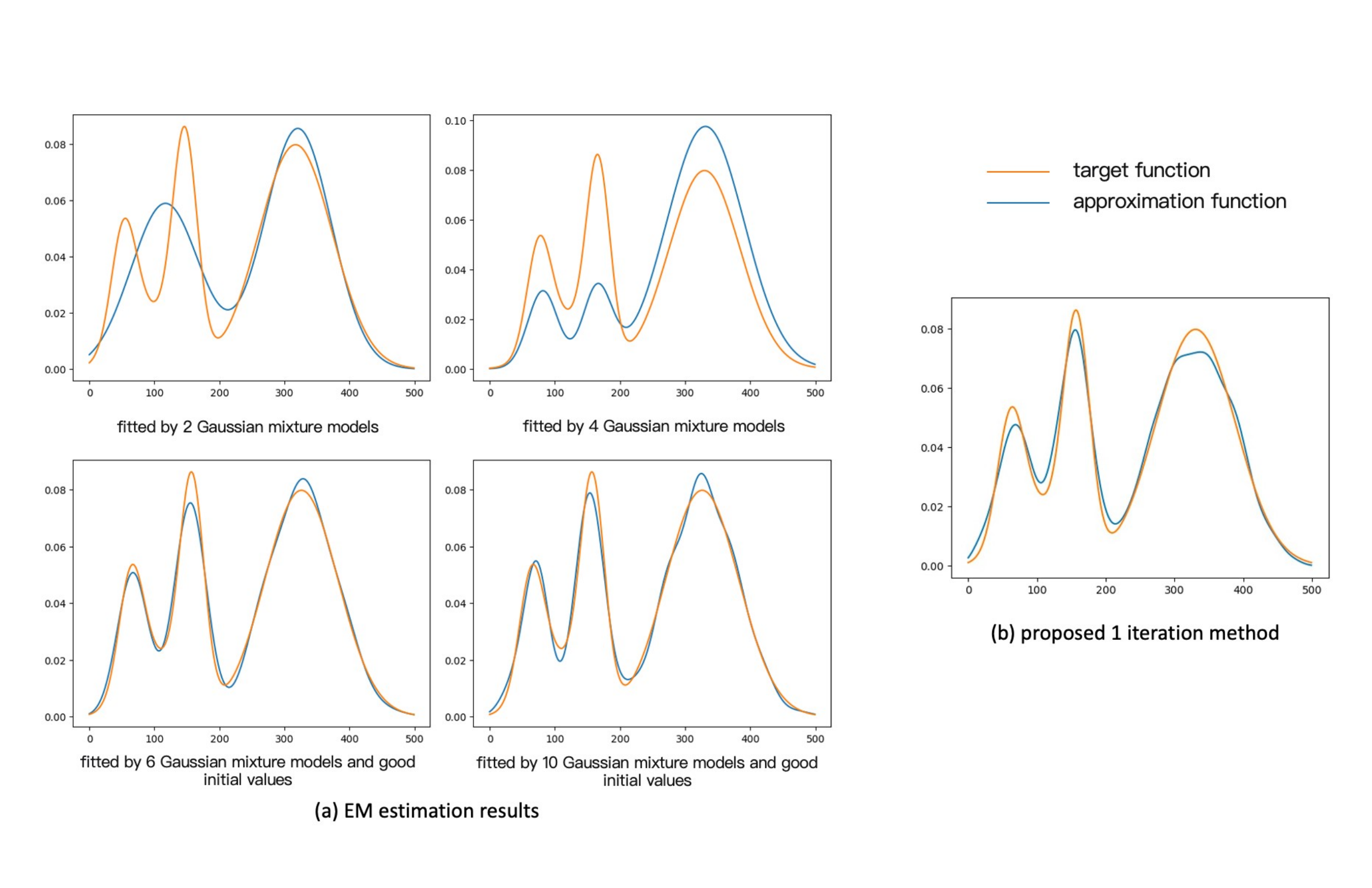}
	\caption{EM VS Proposed Method: Fitting Gaussian Mixture Distribution}
	\label{exp}
\end{figure}

\begin{figure}[ht]
	\centering
		\includegraphics[scale=0.8]{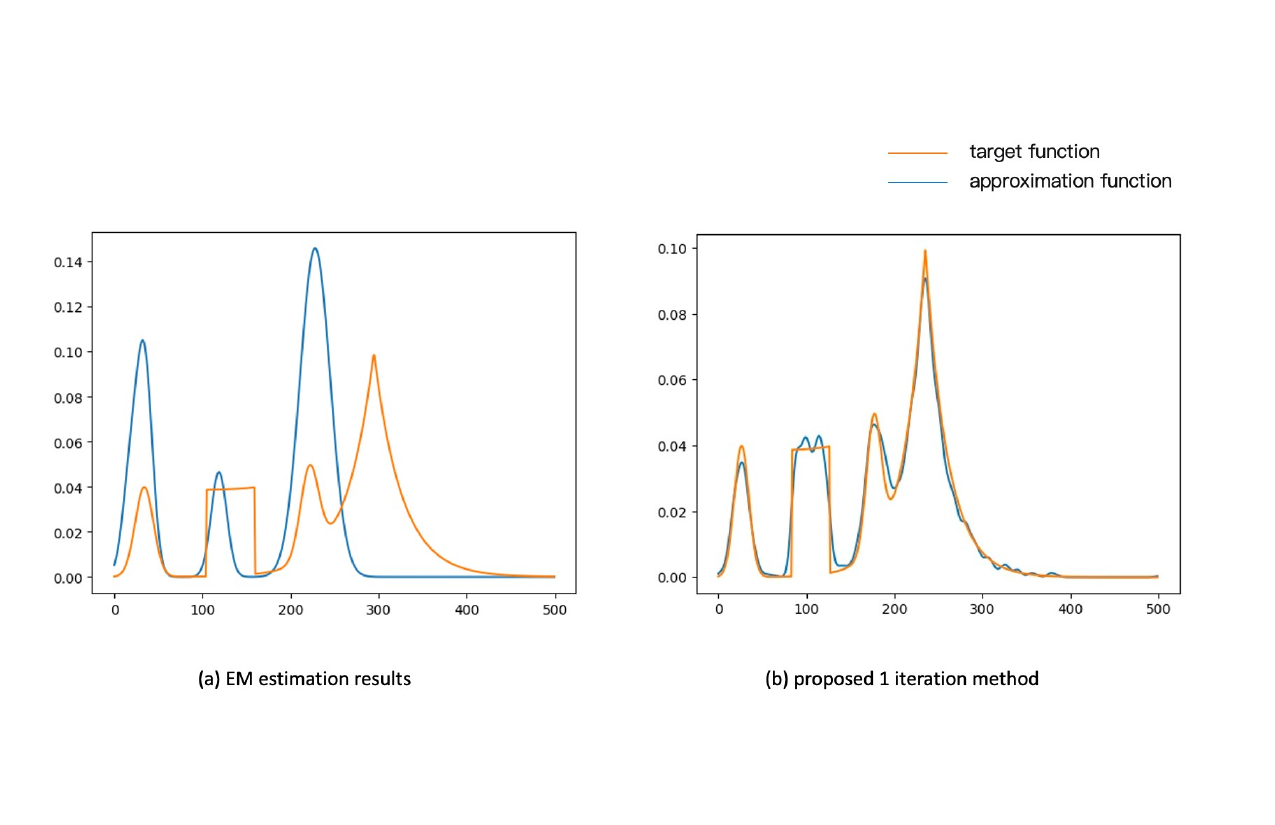}
	\caption{EM VS Proposed Method: Fitting Mixture Multiple Types of Continuous Distribution}
	\label{pic15}
\end{figure}

\begin{figure}[ht]
	\centering
		\includegraphics[scale=0.6]{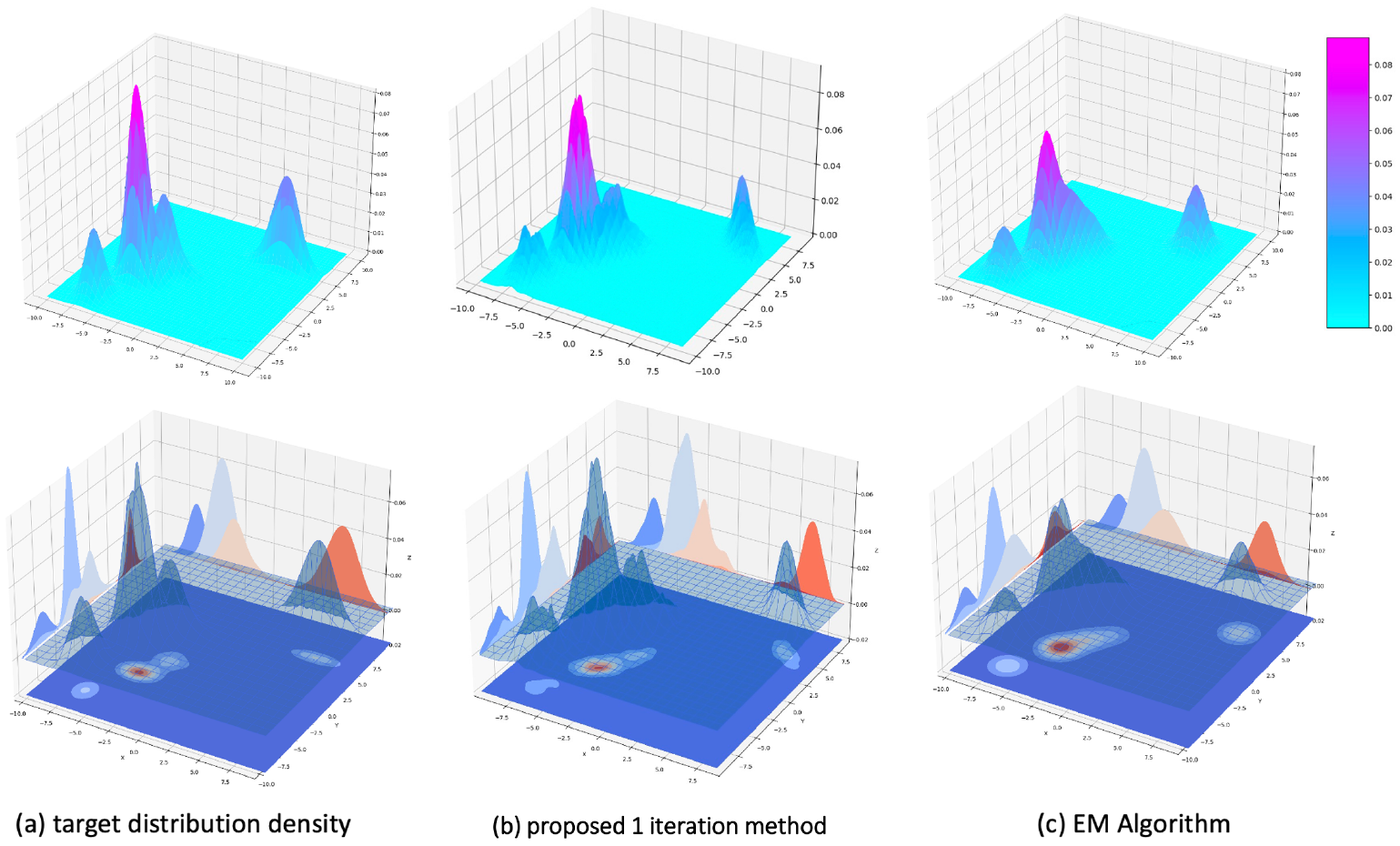}
	\caption{EM VS Proposed Method: 2 Dimension.}
	\label{pic3}
\end{figure}

In Figure.\ref{exp}, the subplot in (b) demonstrate the approximation using our one-iteration method, whereas the four subplots in (a) shows the approximations using the EM algorithm with different initial settings. The target density in this numerical example is produced by mixing four normal distributions. In subplot (a) shows 4 EM estimations by different initialisation and parameters selection. 
2 Gaussian mixtures are used in the top left plot, 4 in the top right, 6 in the bottom left, and 10 in the bottom right. Top left demonstrates that a significant amount of information will be lost if we use Gaussian units that are fewer than all of the peaks present in the target density. 
EM algorithm is initialization-sensitive as well. We show this in the top right plot, where we initialize parameters randomly across data domain. The learning result is stuck in local and fails to find the appropriate direction for optimization\cite{jin2016local}. A appropriate initialization and the proper number of Gaussian units are used to estimate the lower two subplots in (a). It proves that adding more Gaussian units to an EM algorithm does not enhance performance as long as there exist Gaussian units that could cover the target modality. In subplot (b), it demonstrates that our approach produces outcomes that are comparable to the best outcomes that EM was capable of. Our technique has the advantage that it is unaffected by the initialization or number of Gaussian units. Target distribution in Figure.\ref{exp} is a relatively simple cases. When target distributions are more complex for instance a mixture of normal distribution, uniform distribution and Laplace distribution, the differences between the two methods are more stark. Figure.\ref{pic15} shows our results. If the parameters of EM are not set up carefully, it fails to capture the target density modality, whereas our method easily achieves a good estimation that is able to capture most of the modality features. This is one of the main benefits of our algorithm: it is robust and not sensitive to initialization. In the 2-dimensional scenario shown in Figure.\ref{pic3}, Our method yields superior results and allows us to capture the majority of the target density's detail without having to worry about initialization. Comparing to EM algorithm, initialization of means, variances and covariance of EM algorithms are extremely demanding. 

\subsection{Numerical Comparison-IPE}
Due to the convergence of our algorithm, the iterative update of $\pi_n$ does not change after the first step without updating the means and variances, so we only need 1 iteration to obtain the optimal solution. 

\begin{table}[ht]
    \centering
    \caption{ results of two algorithm.}
	\label{table41}
	\resizebox{1\textwidth}{!}{
	\begin{tabular}{lcccc}
	\hline
	algorithm & Gaussian  Unit & Initialize $\pi$ & Iteration & IPE(average 50 trail)  \\       
        \hline 
	Proposed 
 1-iteration method	& 200 & Evenly across & 1 & 0.18358802133136326 \\ 
	EM	& 200 & Evenly across & 5 & 0.22858647422658074  \\ 
        EM	& 50 & Evenly across & 5 & 0.2303729611966345  \\ 
        EM	& 10 & Evenly across & 5 &0.2851425355507268  \\
        EM	& 2 & Evenly across & 5 &0.7026589181688097  \\
        \hline
	\end{tabular}
	}
\end{table}

Table\ref{table41} demonstrates that our new approach can produce a lesser IPE error and a better match for the data. Using more than 50 fitting tests of randomly generated distributions, we calculate the mean IPE error for each method. We randomly generate 1 dimensional mixture distributions with normal distributions and uniform distributions as target distribution. All target distributions are at least have 6 component distributions. Each target distribution sample 2000 data points as dataset for learning GMMs. In our previous work\cite{lu2023efficient}, our experiments shows that 200 units of GMM perform good effectiveness and efficiency. In this test, we believe that over 200 units are not necessary. 

In summary, our algorithm carrying following benefit compare to EM algorithm:
\begin{itemize}
    \item 1. Simplicity: We do not need to think about the initial value selection issue because our model is not initialization-sensitive. The conventional EM algorithm depends heavily on its initial value choice; if the initial value is chosen at random, the process will either fail or find the incorrect local optimal solution. The calculating process for this proposed 1 iteration algorithm is straightforward, and the initialization of our model is straightforward and same for every situation.
    \item 2. Generality: We consider our model to be a general model. The reason is that we experimentally shows our method can approximates a large set of distributions and perform better than EM. 
    \item 3.Robust: Theoretically, we have evidence that the likelihood in our one-iteration method will be better, and the experiment shows that the IPE loss is less than EM. It's important to note that we find our approach to be more stable than the EM algorithm during the training process. When computing the variance using (\ref{com}), it is possible to have a result of 0. A Gaussian density function cannot have zero variance, hence zero variance will prevent the algorithm from operating. This will make it not stable to run the EM algorithm. But this did not occur in our method. 
\end{itemize}

\section{Neural Network and GMM Based Generator}
Latent variables in Autoencoder play crucial role in machine learning, \cite{hinton1993autoencoders,schwenk1997training,lecun1998mnist,goodfellow2020generative,li2017adversarial}. There are many different ways to manipulate latent variables. Variational Autoencoder \cite{kingma2014auto} maps the input variable to a latent space that corresponds to the parameters of a variational distribution. Generative adversarial networks \cite{goodfellow2020generative} use noise variables as input space. Latter on \cite{karras2019style}
embeds the input latent code into an intermediate latent space. VQ-VAE \cite{razavi2019generating, van2017neural} applies the vector quantization method to learn discrete latent representation. Latent diffusion models \cite{rombach2022high} use diffusion processes in pairs with VQ-VAE produce outstanding text to image generation. All these excellent studies point out that latent variables are one of the most important issues for achieving good generation in generative models. In Kolouri \cite{kolouri2018sliced}, they trained an adversarial deep convolutional autoencoder, then learned GMM for latent variable distribution to generate random samples in the embedding space. GMM samples were passed through the decoder to generate synthetic images. We carried out the same experiments, and our learning method shows the same generational quality\cite{lu2023efficient}. Non-Gaussian diffusion model is investigated in \cite{nachmani2021non}. Together with VQ-VAE and latent diffusion model, we have certain degree of evidence that the distribution of latent variables are likely to be non-Gaussians. Insipred by VQ-VAE together will diffusion model and style-GAN, we present a GMM embedding technique to manipulate the latent space's attributes.

\subsection{Cardioid and Inverse problem}

\begin{figure}[ht]
	\centering
		\includegraphics[scale=0.5]{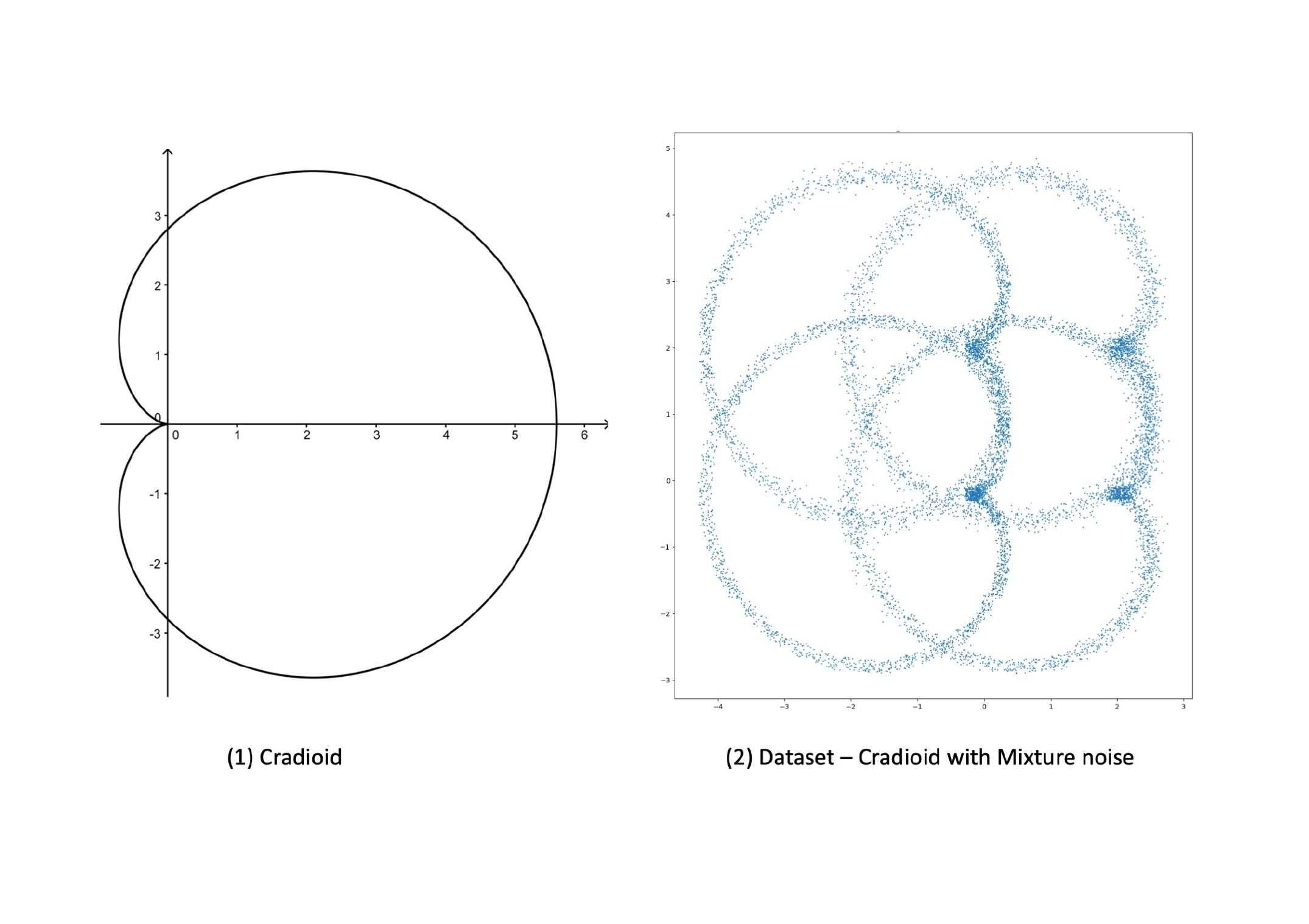}
	\caption{Cardioid}
	\label{pic4}
\end{figure}

With the concept of GMM expansion, we can directly utilize neural networks to map any distribution within a multivariate Gaussian distribution with zero means and an identity covariance matrix. We discovered this by learning a cardioid synthetic dataset. For every point of x in the cardioid, it has 2 values of y and vice versa. This is why inverse problems occur while the inverse function of $f^{-1}\left(y\right)$ is not unique. In Figure.\ref{pic4}, plot 1 is the classic cardioid function. Figure.\ref{pic4} plot 2 is our target dataset. Instead of a simple Gaussian noise, we give both x and y a mixture distribution, where the mixture distribution is constructed by a Gaussian distribution with mean 2.0 and 0.1 variance and a Uniform distribution with [-0.3,-0.1]. These two distributions are mixed with a probability of 0.5.

Classic MLP approach to find the shortest distance of x map to y will have a difficult time handling this problem unless carefully pair each input x with 2 output y. Even though we have carefully paired each input x with 2 outputs, by adding noise into the data, this data input-output becomes difficult to match. Also, the information of the variance will lost. GMM is one of the best solutions for this problem. A TensorFlow-probability example shows that changing the output layer into GMM layers and optimizing it with log-likelihood could solve this problem easily. Based on our proposed 1 iteration GMM learning method, we can change to model output have make a great improvement in fitting.

\begin{figure}[ht]
	\centering
		\includegraphics[scale=0.65]{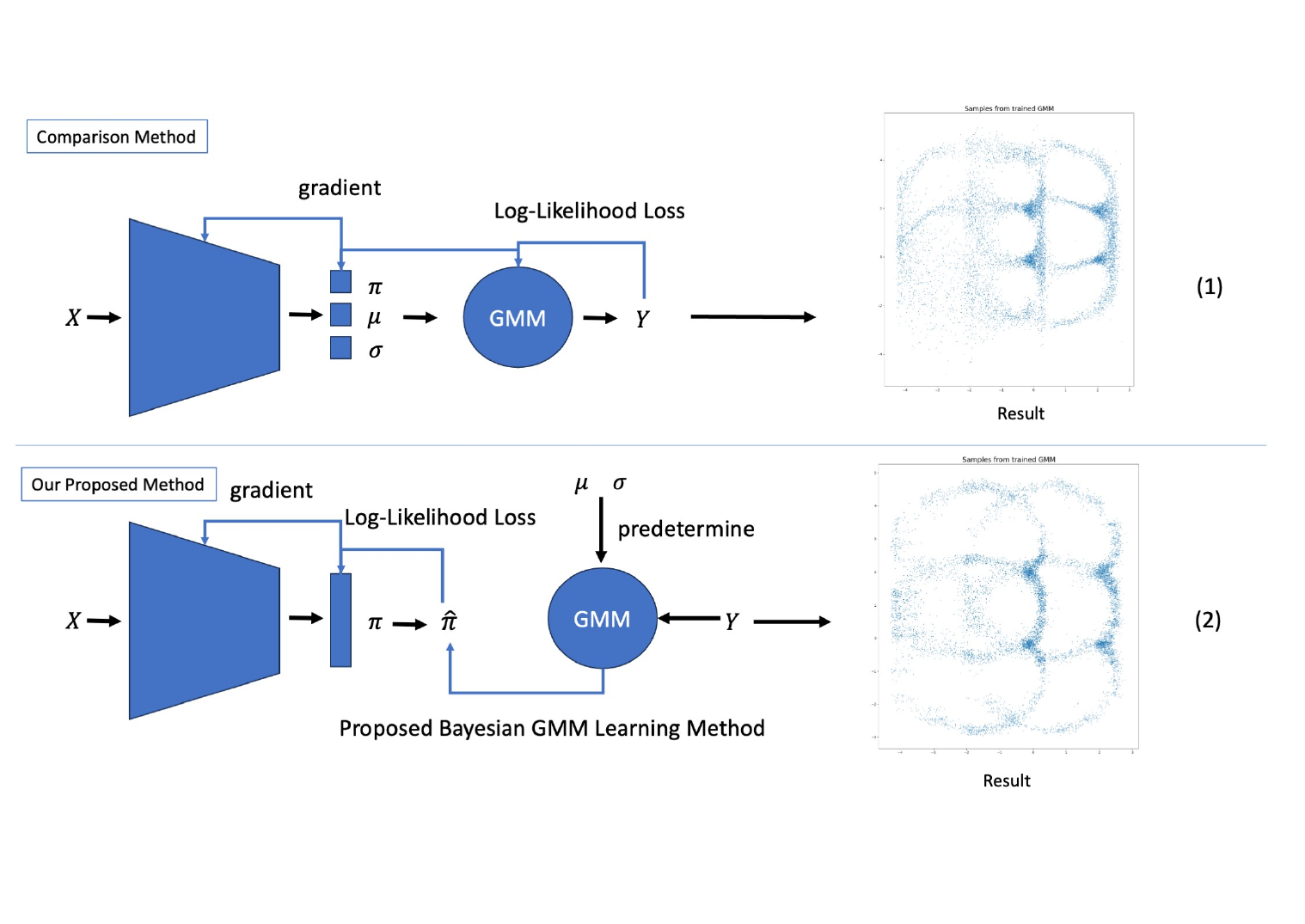}
	\caption{Model Comparison}
	\label{pic5}
\end{figure}

\begin{figure}[ht]
	\centering
		\includegraphics[scale=0.4]{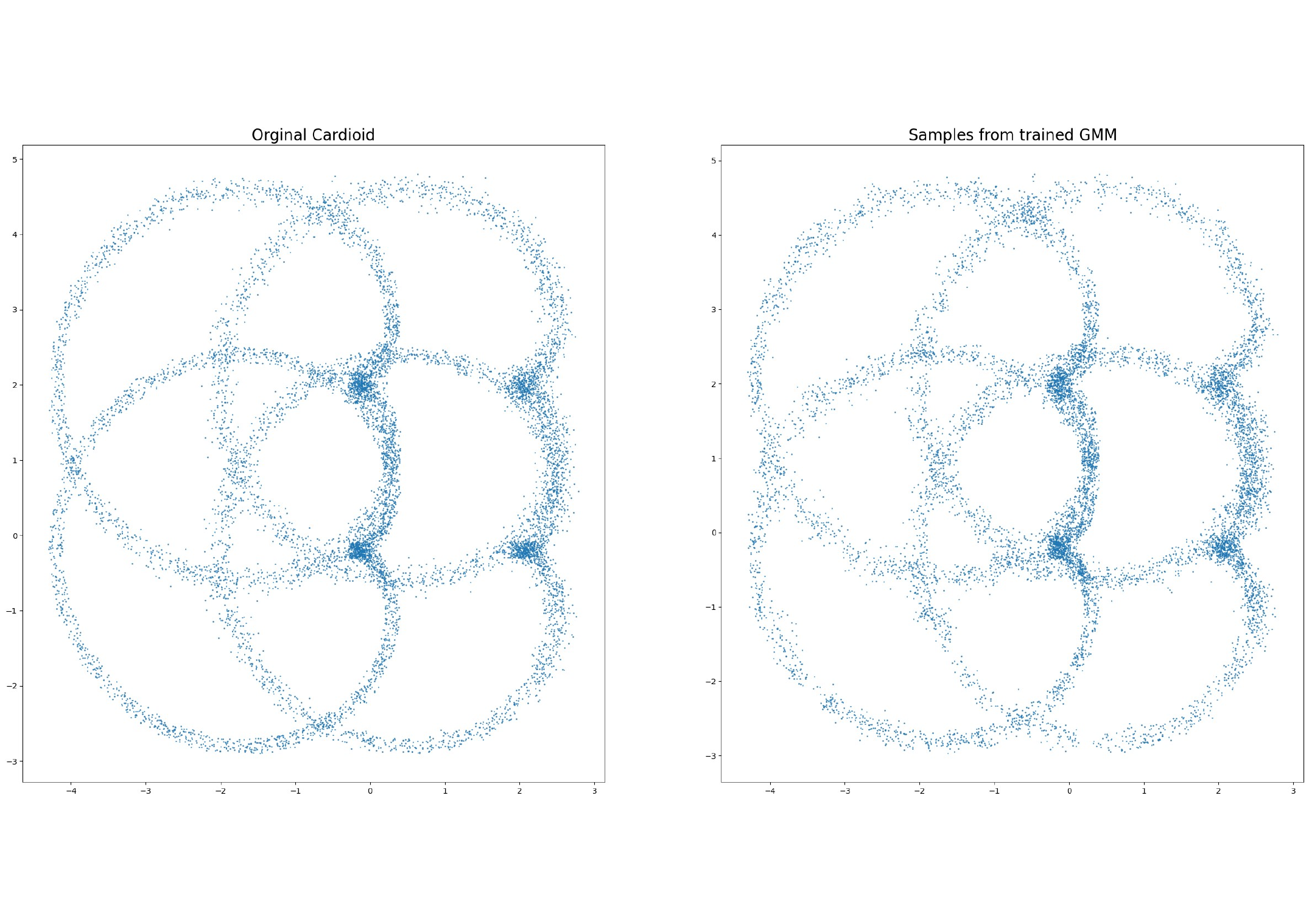}
	\caption{(left)Original Data Distribution. (Right)Approximation with proposed method and 2 Feed Forward Layers}
	\label{pic6}
\end{figure}

Figure.\ref{pic5} summarizes our methodology, outcomes, and head-to-head comparison with the conventional approach. Both neural networks are trained over 200 iterations using a single hidden feedforward network made up of 100 linear units using relu activation. The outcome demonstrates that our approach captures the overall structure more accurately than the conventional way. The data we sampled from learned GMM in Figure.\ref{pic6} demonstrates that our outcome is remarkably close to the original dataset after adding one more hidden layer. This shows the potential of our method, and the GMM expansion concept could improve the accuracy of neural networks. Additionally, latent variable embedding can be done using our method. The following sections provide examples of the technique.

\subsection{GMM based generator}
The Gaussian distribution is without a doubt the most popular distribution in machine learning due to reliable central limit theorem and well-studied features. From variational Autoencoder, GAN, to the latest diffusion model in neural networks and many other models like the Gaussian Process. However, some recent advancements in neural networks are moving away from the Gaussian. Latent diffusion model \cite{rombach2022high}, VQ-VAE\cite{van2017neural} and Style-GAN\cite{karras2019style} are great examples. Both of them provide a significant way to manipulate latent variables and produce a huge impact on image generation. All these studies show that the distribution of latent space is very unlikely to be Gaussian. The latent diffusion model diffuses latent $Z\rightarrow W\sim N(0,1)\rightarrow Z$. Style-Gan\cite{karras2019style} map latent $Z$ to $W$ controls the generator through adaptive instance normalization (AdaIN) at each convolution layer. Gaussian noise is added after each convolution.

All these studies point to the direction that latent space or embedding space is not Gaussian. Most studies aim for a latent space that consists of linear subspace, each of which controls one factor of variation. Our study focuses on directly using GMM to learn subspace distribution. 

\begin{gather}\label{eq32}
    \text{Decoder}(z)=y+\epsilon,\\
    Z\sim \text{Gaussian Mixture Distribution}(\overrightarrow{\pi},\overrightarrow{\mu},\overrightarrow{\Sigma})
\end{gather}

In (\ref{eq32}), the embedding space $Z$ is not linear. Each sample of $Z$ represents a unique mapping from $z$ to $y$. Instead of linearly controlling each parameter in $z$ for a factor of variation, we could control in $z$ by random sampling or based on the Gaussian Mixture Distribution.

\subsubsection{Embedding with GMM vs Nomral vs GMM + Normal}

To determine whether Gaussian mixture distribution is preferable to a standard Gaussian, we examine 3 different embedding types. The variable $Z$ is made up of random samples that are fed straight into the original GAN. The generator is built by transpose convolution layers with input dimension $[1x1x100]$. Convolutional layer transposition with an input dimension of $[1x1x100]$ is used to create the generator. These studies make use of minimalist datasets, with mean square error loss as the loss function. We want to directly evaluate embedding efficiency by randomly mapping images into randomly chosen latent samples, therefore the discriminator is not being employed.

\begin{gather}\label{eq34}
    \begin{bmatrix}z_0 \\.\\.\\.\\z_n \end{bmatrix}=Z,\\
    (1)z_i\sim N(0,1),\\
    (2)z_i\sim GMM(\theta_i),\\
    (3)[z_0...z_{n/2}]\sim N(0,1) \text{ and } [z_{n/2}...z_n]\sim GMM(\theta_i)
\end{gather}

In (\ref{eq34}), we show that how $Z$ is constructed. In (1), $z_i$ are all sampled in Gaussian noise as GAN does. In (2), all $z_i$ are sampled by randomly generated Gaussian mixture distributions which have at least 2 peaks. In (3), the first half of the $z_i$ follows Gaussian distribution and the second half of the $z_i$ are from Gaussian mixture distributions.

\begin{table}[ht]
    \centering
    \caption{ Comparison of sampling $Z$ with 3 different method }
	\label{table2}
	\resizebox{1\textwidth}{!}{
	\begin{tabular}{lcccc}
	\hline
	Distribution & Numbers of $z_i$  & Epochs & Average Loss & Standard deviation of Loss  \\
        \hline
	Normal	& 100 & 50 & 0.09433 & 0.00154 \\
	Gaussian Mixture	& 100 & 50 & 0.09190 & 0.00068  \\
        Gaussian Mixture and Normal	& 100 & 50 & 0.09055 & 0.00090 \\
        \hline
	\end{tabular}
	}
\end{table}

The outcomes of our experiment are shown in Table 2. We train a generator for 50 epochs as a single trail using a random sample of $Z$. We train 30 times for each type of distribution and compute the mean and standard deviation of the loss.  It is obvious that by utilizing a Gaussian Mixture rather than a normal distribution, we have smaller loss and a smaller standard deviation. The best outcome is achieved when using a Gaussian mixing distribution along with a normal distribution.  The standard deviation is better than the normal distribution while slightly greater than the Gaussian Mixture distribution. This supports the argument that we made for the benefits of the Gaussian mixed distribution of latent variables or embedding for neural networks. It also explains why non-Gaussian approaches are taken by the latent diffusion model, style-GAN, and VQ-VAE.

\subsubsection{Embedding with GMM and image generation}

\begin{figure}[ht]
	\centering
		\includegraphics[scale=0.8]{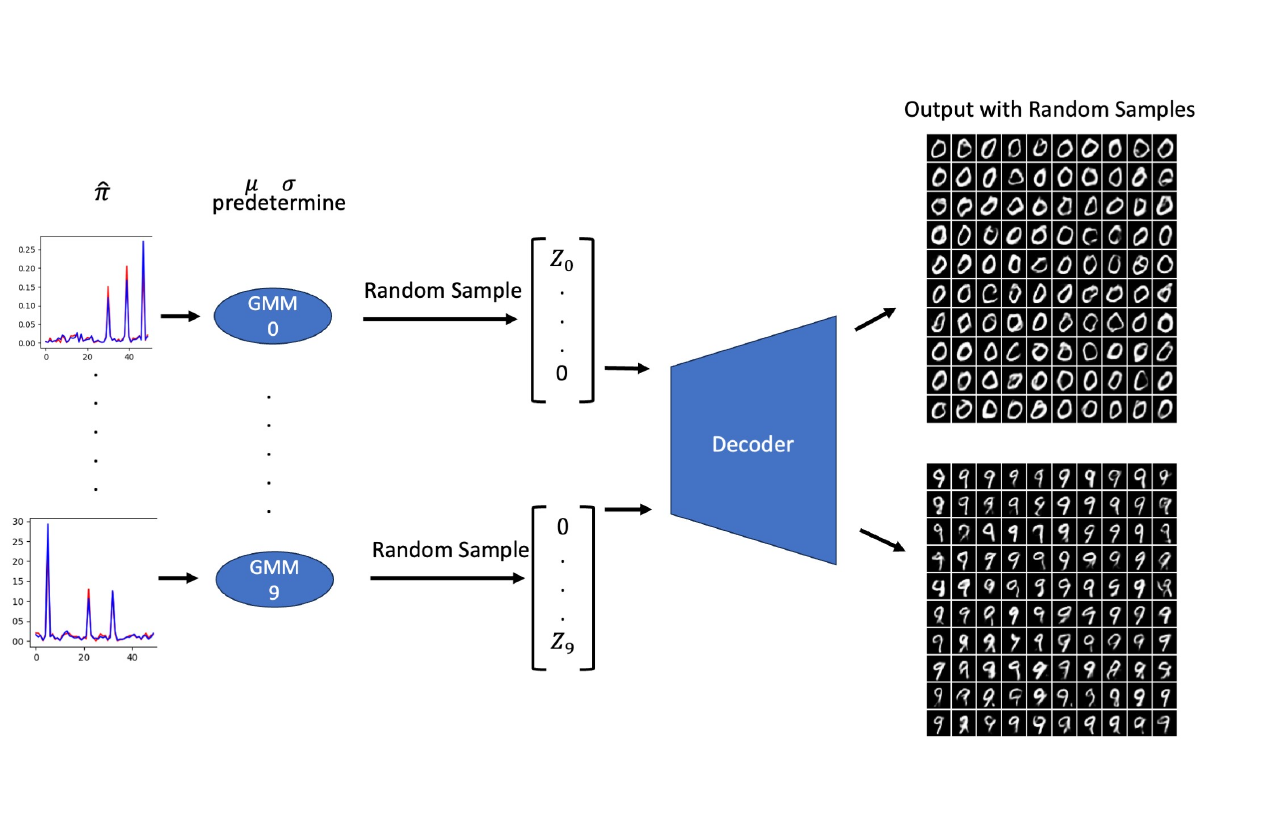}
	\caption{GMM Base Generator}
	\label{pic7}
\end{figure}

In this section we examine whether a neural network can produce meaningful images based on latent variables that sampled from Gaussian mixture distributions. Our experiments using the Minist dataset and our embedding method are shown in Fig.\ref{pic7}.

 $ Z=[Z_0,Z_1,Z_2..Z_{10}]$ is divided into data classes. Each $Z_i$ represents a digit and 20 units for each $Z_i$. Each unit in $Z_i$ are random samples drawn from a mixture distribution. Input's $Z_i$ for unrelated classes will be set to zeros. All Gaussian mixture distributions have at least two peaks.  Depending on the number of peaks, each $Z_i$ have at least $2^{20}$ unique combination for embedding. Mean square error loss is applied. 
 
 Appendix\ref{ap1} shows random samples from trained neural network. Each class has a clear feature construction and separation between each class is clear. This experiment demonstrates that neural networks can adapt to the GMM latent space and produce meaningful generation regardless the form of Gaussian Mixture distribution. It also suggests that we can benefit from random sampling for feature variations. 

\section{Conclusions and Future Work}

Based on both our results and previous work\cite{lu2023efficient}, it is becoming clear that the idea of GMM expansion and our 1 iteration learning algorithm can be a reliable method to learn any distributions. It is better than EM algorithm in terms of simplicity and accuracy. Theoretical proof and experiments shows that our algorithm are reliable and robust.

In the application with neural network, a mixture uncertainty cardioid example shows that the idea of GMM expansion are more suitable for neural network. In previous work we demonstrate GMM can learn latent distribution in encoder decoder architecture. In this work, we further investigate Gaussian mixture distribution embedding for simple generator architecture. GMM stand the test for learning the latent space as well as used as embedding. We expect GMM for directly define intermediate latent space or any GMM based embedding will be able to develop quality large model which we can benefit from random sampling for feature variation. It provide interesting development for future work.

\appendix

\begin{figure}[ht]\label{ap1}
	\centering
		\includegraphics[scale=0.6]{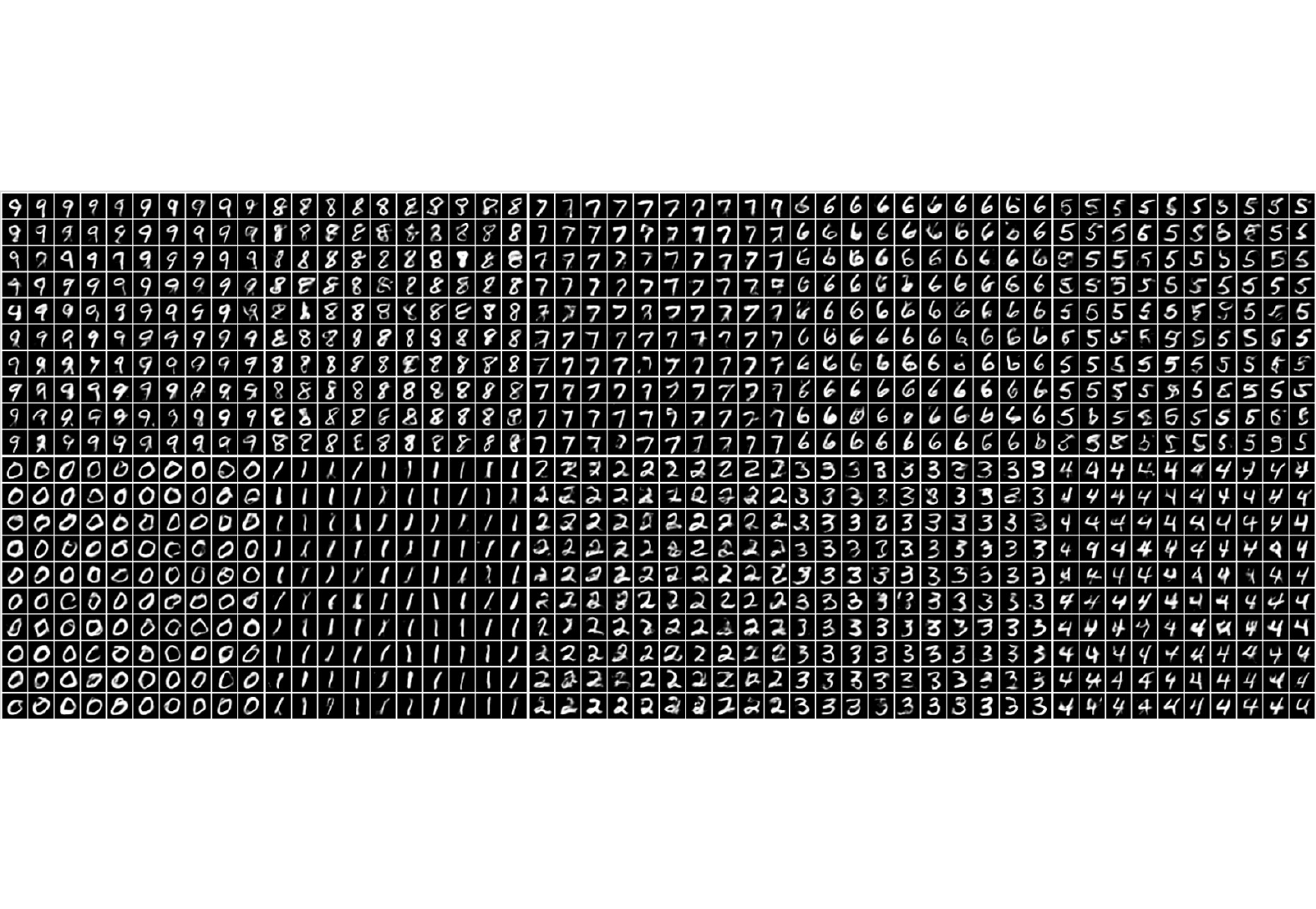}
	\caption{Image Generation with Random Samples from GMM}
	\label{pic17}
\end{figure}



\end{document}